\documentclass[conference]{IEEEtran}
\IEEEoverridecommandlockouts
\usepackage{cite}
\usepackage{amsmath,amssymb,amsfonts}
\usepackage{algorithmic}
\usepackage{graphicx}
\graphicspath{{./fig/}}
\usepackage{textcomp}
\usepackage{xcolor}
\usepackage{siunitx}    
\usepackage{url}
\usepackage{bm}
\usepackage{acronym}
\usepackage{mathtools}
\usepackage{booktabs}
\usepackage{multirow}
\usepackage{dsfont}
\usepackage{tikz}
\usepackage{adjustbox}
\usepackage{stfloats}
\usetikzlibrary{arrows, positioning, calc}

\def\nom{\mathrm{nom}}
\def\mppi{\mathrm{MPPI}}
\def\Lone{\mathcal{L}_1}

\def\BibTeX{{\rm B\kern-.05em{\sc i\kern-.025em b}\kern-.08em
    T\kern-.1667em\lower.7ex\hbox{E}\kern-.125emX}}
\begin{document}

\title{\LARGE \bf L1-MPPI: L1 Adaptive Model Predictive Path Integral\\for Agile UAV Control}


\author{Luk\' a\v{s} Kotek, Ond\v{r}ej Proch\' azka, Voj\v{e}ch Von\' asek, Martin Saska and Robert P\v{e}ni\v{c}ka
\thanks{The authors are with the Multi-robot Systems Group, Faculty of Electrical
Engineering, Czech Technical University in Prague, Czech Republic (\protect\url{http://mrs.felk.cvut.cz/}). 
This work has been supported by the Czech Science Foundation (GAČR) under research project No. 23-06162M, by the European Union under the project Robotics and advanced industrial production (reg. no. CZ.02.01.01/00/22\_008/0004590), and by CTU grant no SGS26/077/OHK3/1T/13.
}}

\maketitle

\begin{abstract}
This work proposes the $\bm{\mathcal{L}_1}$ Adaptive Model Predictive Path Integral ($\bm{\mathcal{L}_1}$-MPPI).
It cascades $\bm{\mathcal{L}_1}$ adaptive control with the Model Predictive Path Integral (MPPI) to improve tracking of high-speed UAV trajectories.
Thanks to the $\bm{\mathcal{L}_1}$ augmentation, the tracking remains accurate even under model uncertainties and external disturbances, such as an additional payload or a mismatch in the modeled aerodynamic drag.
In contrast to existing MPPI approaches for UAV control that do not explicitly model aerodynamic effects, varying payloads, and typically neglect the dynamics of low-level motor controllers, our 
$\bm{\mathcal{L}_1}$-MPPI approach enhances the dynamic model used in the MPPI by incorporating the low-level flight controller and motor dynamics, as well as an iterative mixing scheme that reflects the approach of the low-level controller.
The proposed method demonstrates improved tracking performance in both simulation and the real world, even when the UAV is subjected to an unknown payload. 
In flight with \textbf{\SI{35}{\percent}} mass increase, our approach lowers the RMSE by \textbf{\SI{58.61}{\percent}} with respect to plain MPPI.
Compared to the same MPPI using an online mass estimator in place of the $\bm{\mathcal{L}_1}$ augmentation, the RMSE is lower by \textbf{\SI{38.59}{\percent}}.
During the real-world experiments the UAV reaches speeds up to \textbf{\SI{13.50}{\meter\per\second}} and accelerations up to \textbf{\SI{2.5}{\g}}.
\end{abstract}



\vspace{-2em}
\section*{Supplementary Material}
{\footnotesize
\vspace{-0.3em}
\noindent \textbf{Video:} \url{https://youtu.be/ljqrR1NBq2w}
\vspace{-0.7em}
}

\section{Introduction} \label{sec:introduction}
Unmanned Aerial Vehicles (UAVs) are increasingly adopted across a wide range of applications, including package delivery~\cite{betti_sorbelli_uav-based_2024}, infrastructure monitoring~\cite{burri_aerial_2012}, the digitization of historical monuments~\cite{petracek_new_2024}, and time-critical search-and-rescue missions~\cite{faessler_autonomous_2016}.
Many of these applications benefit from fast and agile flight, which shortens the mission duration.
Such flight is possible thanks to high-speed cameras and other advanced onboard sensors.
During aggressive maneuvers, however, the nominal dynamics deviate more from the modeled system dynamics.
UAVs must therefore track agile trajectories accurately even under model uncertainties and external disturbances, such as uncertain aerodynamic drag, varying payloads, and wind gusts.
If left uncompensated, the resulting deviations from the planned trajectory increase the likelihood of collisions and can lead to loss of stability or to mission failure by missing a~crucial waypoint or sensor viewpoint.


\begin{figure}[!t]
    \centering
    \includegraphics[width=\columnwidth]{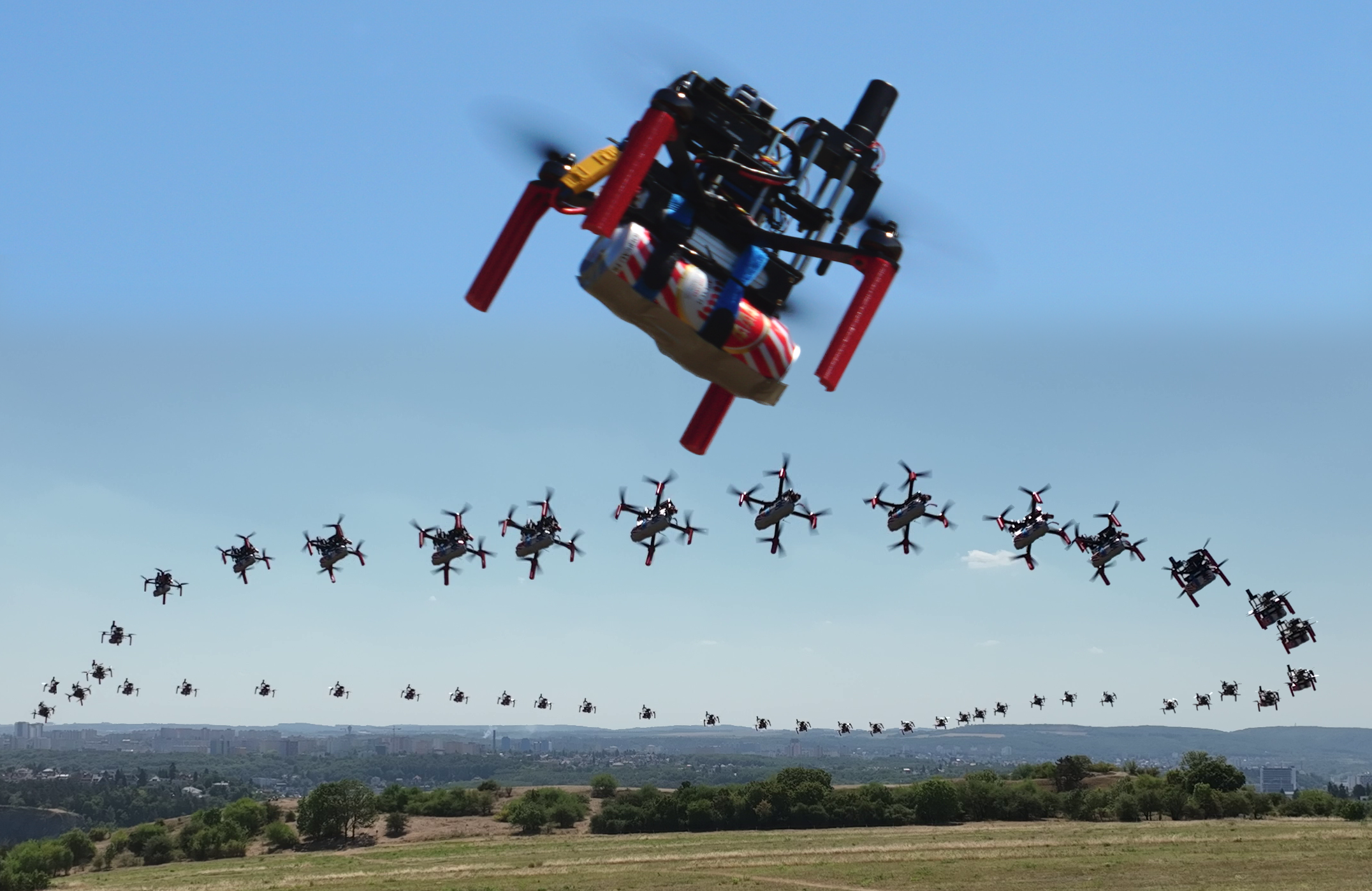}\vspace{-0.4em}
    \caption{Real-world agile flight of the UAV controlled by the proposed $\mathcal{L}_1$-MPPI at speeds up to \SI{13.00}{\meter\per\s} and accelerations reaching \SI{2.5}{\g}. 
    The UAV carries a payload of unknown weight, and the $\mathcal{L}_1$ augmentation compensates for the resulting model mismatch online. 
    The composite image shows the flown trajectory, with a detail of the UAV with the attached payload.}
    \label{fig:real_world}
\end{figure}

Recently, in agile UAV control, Model Predictive Controllers (MPCs) have shown remarkable performance in tracking partially infeasible trajectories~\cite{sun_comparative_2022, gupta_lol-nmpc_2025}. 
However, an inaccurate dynamic model may lead to increased tracking errors and even system instability~\cite{mo_nonlinear_2019}. 
Existing research has therefore primarily focused on improving the UAV dynamic model.
Data-driven methods, such as~\cite{torrente_data-driven_2021}, learn accurate models from data, but the learned models tend to overfit and are difficult to update online.
Another source of prediction error is the hierarchical control architecture commonly used on UAV platforms~\cite{baca_mrs_2021}.
The high-level controller sends commands to a~low-level controller, whose dynamics are usually neglected in the dynamic model.
This leads to trajectory-tracking errors, especially during aggressive maneuvers.
Approaches that model the low-level and inner-loop controllers~\cite{gupta_lol-nmpc_2025, sun_comparative_2022} demonstrate a~significant reduction in the trajectory error.
Still, no dynamic model captures the uncertainties and disturbances described above, since they vary in time and are unknown beforehand.
Adaptive control approaches estimate and compensate for them online, thereby improving robustness, but they require complex adaptation laws~\cite{mo_nonlinear_2019}.
The authors of~\cite{hanover_performance_2022} therefore cascade Nonlinear MPC (NMPC) with an $\mathcal{L}_1$ adaptive controller.
This preserves the trajectory-tracking capabilities of NMPC and adds online compensation of external disturbances and model uncertainties.

This paper proposes a~Model Predictive Path Integral (MPPI)-based controller combined with an $\mathcal{L}_1$ adaptive controller (also referred to as $\mathcal{L}_1$ augmentation). 
MPPI uses a~Graphics Processing Unit (GPU) to compute hundreds of candidate trajectories (rollouts) in parallel in each iteration~\cite{williams_model_2017}. 
In~\cite{minarik_model_2024, pochobradsky_geometric_2026} MPPI demonstrated the ability to track trajectories while avoiding obstacles in an unknown environment. 
Unlike these works, our focus is not on obstacle avoidance but on improving trajectory tracking under model uncertainties.
Similarly to~\cite{gupta_lol-nmpc_2025}, we incorporate the low-level controller and motor dynamics into the MPPI's dynamic model.
We additionally account for communication delays between the high-level controller, flight controller, and state estimator.
On top of this model, the $\mathcal{L}_1$ adaptive controller estimates the remaining model uncertainties online and compensates for them in real time.
As a result, the proposed framework enables agile flight in the presence of model uncertainties and external disturbances. 
To the best of our knowledge, the proposed $\mathcal{L}_1$-MPPI is the first adaptive MPPI framework that models the low-level controller, the motor dynamics, and the communication delays, and the first adaptive MPPI demonstrated in real-world agile flight.
In contrast to~\cite{pravitra_l1-adaptive_2020}, our implementation runs fully onboard at \SI{100}{\hertz}.

The proposed $\mathcal{L}_1$-MPPI controller is evaluated in both simulation and real-world experiments assessing adaptive capabilities during agile flights.
We compare the proposed controller with a plain MPPI controller and with our MPPI approach that employs online mass estimation instead of $\mathcal{L}_1$ augmentation. 
The $\mathcal{L}_1$ augmentation alone, compared to the same controller without it, has only a minor effect on the positional RMSE under nominal conditions, but reduces it by~\SI{51.11}{\percent} in the presence of an unknown payload. 
The complete \mbox{$\mathcal{L}_1$-MPPI}, which additionally models the low-level controller and motor dynamics, reduces the positional RMSE by \SI{24.78}{\percent} with respect to plain MPPI under nominal conditions and by \SI{44.37}{\percent} in the presence of uncertainties. 
Moreover, the proposed approach demonstrates its adaptability in real-world agile flight experiments with an attached payload (see Fig.~\ref{fig:real_world}).


\section{Related Works}
This section reviews the three topics the proposed method builds upon: the UAV dynamic model used by predictive controllers, adaptive controllers that compensate online for what such a model cannot capture, and MPPI-based UAV control.


To operate at the platform's performance limits, the dynamic model used in predictive controllers must closely represent the UAV's true behavior.
One way to improve the model is to learn it from data, using Gaussian Processes~\cite{torrente_data-driven_2021}, neural ordinary differential equations~\cite{chee_knode-mpc_2022}, and physics-informed neural networks~\cite{sanyal_ramp-net_2023}.
However, their reliance on offline training and off-board computation limits adaptation to changing dynamics.

Other works instead model the neglected effects explicitly.
In~\cite{faessler_differential_2018}, the authors demonstrate how modeling aerodynamic drag can improve the tracking accuracy of a~Differential-Flatness-Based Controller (DFBC) in agile flight.
In~\cite{faessler_thrust_2017}, motor dynamics and iterative motor mixing are incorporated to improve tracking performance and prevent actuator saturation during aggressive maneuvers.
The authors of~\cite{gupta_lol-nmpc_2025} improve tracking performance by incorporating low-level controller dynamics and iterative motor mixing into the NMPC model.
The approach respects the actuator constraints, runs onboard, and is validated in real-world flight.

Despite advances in dynamic modeling, prediction errors arising from model uncertainties and external disturbances remain unavoidable. 
Adaptive control methods address this limitation by compensating for modeling errors online while preserving robust performance~\cite{lavretsky_robust_2024}. 
Several adaptive controllers have been proposed for UAVs~\cite{dydek_adaptive_2013, madebo_robust_2024}. 
However, most have only been demonstrated during hovering or low-speed flight.


A composite Model Reference Adaptive Controller (MRAC), which forces the closed loop to follow the behavior of a chosen reference model, is combined with a baseline trajectory-tracking controller in~\cite{dydek_adaptive_2013}, demonstrating adaptability both in simulation and on a UAV subjected to a sudden thrust reduction. 
However, the controller is executed off-board and is not evaluated on agile trajectories. 
A neural-network-based MRAC is proposed in~\cite{madebo_robust_2024}, but its adaptation rate is insufficient for agile flight, with experiments conducted at speeds of only \SI{2}{\meter\per\s}. 
Mass estimation methods based on Inertial Measurement Unit (IMU) measurements are applied for payload compensation in~\cite{zhou_hybrid_2021}, although they are limited to slowly varying payloads. 
The adaptive controller proposed in~\cite{xie_adaptive_2022} compensates for a broader range of uncertainties, including wind disturbances, but is also evaluated only on low-speed trajectories.


A faster alternative to MRAC is $\mathcal{L}_1$ adaptive control, which decouples the adaptation rate from the robustness of the closed loop by low-pass filtering the estimated uncertainty. 
It has been successfully applied to a variety of aerospace vehicles~\cite{lee_l1_2012, gregory_l1_2009}, including UAVs~\cite{pravitra_l1-adaptive_2020, hanover_performance_2022}. 
In~\cite{pravitra_l1-adaptive_2020}, the authors combine $\mathcal{L}_1$ adaptive control with MPPI, demonstrating improved trajectory tracking under drag and mass uncertainties. 
Their dynamic model treats the UAV as a rigid body driven directly by the commanded inputs, without the low-level controller dynamics and the communication delays.
The approach is also not evaluated against other adaptive controllers and is demonstrated only in a simulated racing environment. 
The authors of~\cite{hanover_performance_2022} combine $\mathcal{L}_1$ adaptive control with NMPC, demonstrating robust trajectory tracking during agile real-world flights under payload variations and other model uncertainties.


Model Predictive Path Integral (MPPI) is a~predictive control approach capable of controlling nonlinear systems~\cite{williams_model_2017, minarik_model_2024, zhai_pa-mppi_2026}. 
As aforementioned, MPPI with an $\mathcal{L}_1$ adaptive controller is used in~\cite{pravitra_l1-adaptive_2020} for UAV control.
In~\cite{minarik_model_2024}, the authors introduce a fully onboard MPPI-based controller for agile trajectory tracking, validated in simulation and real-world experiments, with obstacle avoidance shown only in simulation.
The authors of~\cite{pochobradsky_geometric_2026} extend this work with an $SE(3)$ controller applied to a subset of the generated trajectories, and demonstrate real-world obstacle avoidance using an onboard depth camera.
Perception-Aware MPPI~\cite{zhai_pa-mppi_2026} adds a perception-driven cost that adapts the trajectory online and enables navigation to occluded goals, but reports higher failure rates under wind disturbances due to the lack of adaptive compensation.
None of these MPPI approaches models the low-level controller, the motor dynamics, or the communication delays.

Our proposed $\mathcal{L}_1$-MPPI cascades the $\Lone$ augmentation with MPPI instead of the NMPC used in~\cite{hanover_performance_2022}, which keeps the sampling-based cost of MPPI while targeting the same agile flight regime.
Compared to the $\Lone$-augmented MPPI of~\cite{pravitra_l1-adaptive_2020}, we additionally model the low-level PID controller, the motor dynamics, and the communication delays, and we evaluate the controller onboard in real-world flight rather than in simulation only.
Compared to the MPPI controllers of~\cite{minarik_model_2024, pochobradsky_geometric_2026, zhai_pa-mppi_2026}, which rely on a nominal model without adaptation, we compensate for the remaining model uncertainties and external disturbances online.

\section{Methodology} \label{sec:methodology}

\begin{figure}[!hp]
    \centering
    \begin{adjustbox}{width=0.48\textwidth}
        \tikzset{
  >=stealth',
  block/.style={
    rectangle,
    rounded corners,
    draw=black, 
    very thick,
    minimum width=3.8cm,
    minimum height=1.1cm,
    align=center,
    font=\large
  },
  sum/.style={
    circle,
    draw=black,
    very thick,
    minimum size=0.9cm,
    inner sep=1pt
  },
  arrow/.style={
    ->,
    very thick,
    rounded corners,
    shorten <=2pt,
    shorten >=2pt,
  }
}

\begin{tikzpicture}[node distance=2cm and 2cm, auto, every node/.append style={font=\large}]

\node[block] (quad) {UAV Plant};
\node[block, right=3cm of quad] (mppi) {MPPI Controller};

\node[block, below =3.4cm of quad, minimum width=4.0cm, minimum height=1.3cm] (px4) {PID-Based Low-Level \\ Flight Controller};

\node[block, right=1.6cm of px4, minimum width=3.2cm, minimum height=1.3cm] (mapping) {Thrust-Throttle \\ Mapping};

\node[sum, right=1.6cm of mapping, font=\LARGE] (sum) {+};

\node[block, above=1.2cm of sum] (adaptive) {\(\mathcal{L}_1\) Adaptive\\Controller};

\coordinate (loop) at ($(adaptive.east) + (0.9, 0)$);

\draw[arrow] (quad.east) -- node[above] {\(\mathbf{p}, \mathbf{v}, \mathbf{q}, \bm{\omega}, \bm{\Omega}\)} (mppi.west);

\draw[arrow] 
  ($(quad.east) + (2.0, 0.08)$)
  -- ++(0, -1.5)
  -| ($(adaptive.north) + (-0.7, 0)$);

\draw[arrow] (adaptive.south) -- node[right] {\({F}_{\mathcal{L}1}, \bm{\omega}_{\mathcal{L}1}\)} (sum.north);

\draw[arrow] (sum.west) -- node[above] {\(F_t, \bm{\omega}_c\)} (mapping.east);

\draw[arrow] (mapping.west) -- node[above] {\(t_c, \bm{\omega}_c\)} (px4.east);

\draw[arrow] (px4.north) -- node[right] {\(\bm{r}_c\)} (quad.south);

\draw[arrow] (quad.west) -- ++(-0.8, 0) |- node[left, pos=0.25] {\(\bm{\omega}\)} (px4.west);

\draw[arrow]
  ($(loop |- mppi.east) + (0.3, 0.2)$) node[above] {\(\mathcal{T}_{ref}\)}
  -- ($(mppi.east) + (0, 0.2)$);

\draw[arrow] ($(loop) + (0.08, 0)$) -- (adaptive.east);
\draw[arrow] 
  ($(mppi.east) + (0,-0.2)$) 
  -| node[left, pos=0.64] {\(F_{MPPI}, \bm{\omega}_{MPPI}\)} (loop)
  |- (sum.east);
  
\end{tikzpicture}
    \end{adjustbox}
    \caption{
    The proposed $\Lone$-MPPI control architecture.
    The state estimate $\mathbf{p}, \mathbf{v}, \mathbf{q}, \bm{\omega}, \bm{\Omega}$ is provided to both the MPPI and $\Lone$ adaptive controllers.
    The MPPI controller computes the nominal control input $F_{MPPI}, \bm{\omega}_{MPPI}$ based on the reference trajectory $\mathcal{T}_{ref}$, while the $\Lone$ controller generates the adaptive compensation command $F_{\mathcal{L}1}, \bm{\omega}_{\mathcal{L}1}$.
    The combined command $t_c, \bm{\omega}_c$ is sent to the PID-based low-level flight controller, which computes individual rotor throttle commands $\bm{r}_c$ for the UAV.
    The entire control architecture operates at \SI{100}{\Hz}.}
    \label{fig:control_architecture}
\end{figure}
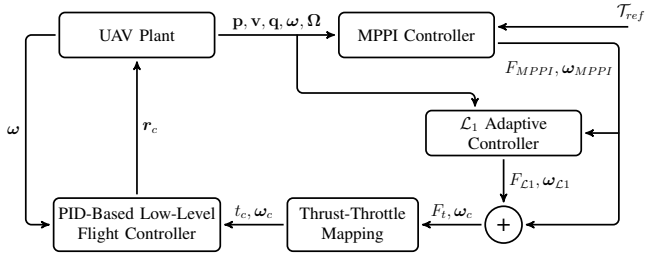

This section describes the building blocks of the proposed MPPI architecture (Fig.~\ref{fig:control_architecture}).
Its core consists of two models, namely the dynamic model of the UAV (Section~\ref{sec:uavmodel}) and the model of its low-level flight controller (Section~\ref{sec:lol_dynamics}).
Our first contribution is to consider both UAV model and its flight controller to achieve precise trajectory following using MPPI (Section~\ref{sec:mppi}).
MPPI provides nominal control (thrust $F_\mppi$ and body-rate $\omega_\mppi$).
Our second contribution is the $\Lone$ adaptive augmentation of MPPI (Section~\ref{sec:l1_adaptive_controller}).
It compensates for model uncertainties and external disturbances (e.g., unknown or changing payload) by providing compensating thrust $F_{\Lone}$ and body-rate $\omega_{\Lone}$.


\subsection{UAV Model}
\label{sec:uavmodel}

The UAV state is described by
position $\mathbf{p} \in \mathbb{R}^3$,
velocity $\mathbf{v} \in \mathbb{R}^3$,
body rate $\bm{\omega} \in \mathbb{R}^3$ and
and the rotation $\mathbf{q}$ represented as a quaternion.
The nominal UAV dynamic model is
\begin{equation}
\label{eq:uav_eqs}
\begin{gathered}
\dot{\mathbf{p}} = \mathbf{v}, \quad
\dot{\mathbf{q}} = \frac{1}{2}\mathbf{q}\odot
\begin{bmatrix}0\\\bm{\omega}\end{bmatrix},
\quad
\dot{\bm{\omega}} =
\mathbf{J}^{-1}
\left(
\bm{\tau}-\bm{\omega}\times\mathbf{J}\bm{\omega}
\right),
\\[0.3em]
\dot{\mathbf{v}} =
\frac{1}{m}\mathbf{R}(\mathbf{q})
\left(
\begin{bmatrix}0\\0\\F_t\end{bmatrix}
-
\mathbf{D}\mathbf{R}^T(\mathbf{q})\mathbf{v}
\right)
+\mathbf{g} \; ,
\end{gathered}
\end{equation}
where $\mathbf{R}(\mathbf{q})$ is the rotation matrix corresponding to $\mathbf{q}$, $\odot$~denotes quaternion multiplication, $\mathbf{J}$ is the inertia matrix, $m$ is the UAV mass, and linear aerodynamic drag is approximated by the diagonal matrix $\mathbf{D}$~\cite{faessler_differential_2018}.
The control inputs of the nominal dynamic model~\eqref{eq:uav_eqs} consist of the collective thrust $F_t$ and body torques $\bm{\tau}=[\tau_x,\tau_y,\tau_z]^T$.

Often, high-level controllers (e.g., trajectory tracking controllers like our MPPI) output collective thrust $F_{t}$ and the body-rate commands $\bm{\omega}_{c}$. 
The desired thrust is then converted into a normalized collective throttle command $t_c$ using a static mapping between the throttle, the produced thrust, and the battery voltage.
The UAV is then controlled using a low-level flight controller that, based on the commands
$t_c$ and $\omega_c$, controls the motors.

The behavior of the low-level flight controller is usually ignored in the literature, yet it influences the behavior of UAVs.
Therefore, in this paper, we propose to also consider the dynamics of the low-level flight controller inside the MPPI controller; these dynamics are described in the next subsection.


\subsection{Dynamics of the Low-level flight Controller} 
\label{sec:lol_dynamics}



The flight controller accepts the collective throttle commands $t_c$ and the body-rate command $\bm{\omega}_c$ and provides control inputs $F_t$ and $\tau$ for the UAV model (\ref{eq:uav_eqs}).

The PID-based flight controller dynamics are modeled as
\begin{equation}
\label{eq:pid_eqs}
    \begin{aligned}
        &\bm{e} = \bm{\omega}_c - \bm{\omega}, \\
        &\dot{\bm{e}_I} = \bm{e}, \\
        &\bm{\tau}_{c} = k_p \bm{e} + k_i \bm{e}_I,
    \end{aligned}
\end{equation}
where $\bm{\tau}_{c}$ represents the target torques generated by the PID controller.
The derivative term is neglected, as the prediction time step $\Delta t$ is too large to accurately capture the transient behavior of the body-rate dynamics, making a finite-difference approximation of $\dot{\bm{e}}$ unreliable.

The normalized allocation matrix $\bm{G}$ is then used to compute individual motor throttle commands from the collective throttle and target torques
\begin{equation}
    \underbrace{\begin{bmatrix}
        r_{c,1} \\
        r_{c,2} \\
        r_{c,3} \\
        r_{c,4}
    \end{bmatrix}}_{\bm{r}_c}
     =
     \underbrace{\begin{bmatrix}
         1 & -0.7071 & -0.7071 & -1.0 \\
         1 & 0.7071 & 0.7071 & -1.0 \\
         1 & 0.7071 & -0.7071 & 1.0 \\
         1 & -0.7071 & 0.7071 & 1.0
     \end{bmatrix}}_{\bm{G}}
     \begin{bmatrix}
         t_c \\
         \bm{\tau}_c
     \end{bmatrix}.
\end{equation}
Using the desired normalized rotor throttles $\bm{r}_c$, the motor dynamics are modeled as
\begin{equation}
    \dot{\bm{r}} = \frac{1}{k_{\mathrm{mot}}}(\bm{r}_c-\bm{r}),
\end{equation}
where $k_{\mathrm{mot}}$ is the time constant of the first-order motor model, and
$r_i \in \bm{r}$ is 
the normalized motor angular velocity
\begin{equation}
\label{eq:motor_speed_to_throttle}
r_i = \Omega_i / \Omega_{\max},
\end{equation}
\noindent
where $\Omega_{\max}$ is the maximum motor angular velocity.

The individual rotor thrusts are obtained as
\begin{equation}
    \label{eq:single_rotor_thrusts}
    \bm{f}=f_{\max}(\bm{r}\otimes\bm{r}),
\end{equation}
where $f_{\max}$ is the maximum thrust of each motor and $\otimes$ denotes element-wise multiplication.

The control inputs of the UAV dynamics in \eqref{eq:uav_eqs} are then computed as
\begin{equation}
    \begin{bmatrix}
        F_t\\
        \bm{\tau}
    \end{bmatrix}
    =
    \bm{\Gamma}\bm{f},
\end{equation}
where $\bm{\Gamma}$ is the allocation matrix mapping
individual actuator commands to the overall force and moment vector
\begin{equation}
    \bm{\Gamma}=
    \begin{bmatrix}
        1 & 1 & 1 & 1\\
        -l/\sqrt{2} & l/\sqrt{2} & l/\sqrt{2} & -l/\sqrt{2}\\
        -l/\sqrt{2} & l/\sqrt{2} & -l/\sqrt{2} & l/\sqrt{2}\\
        -c_{tf} & -c_{tf} & c_{tf} & c_{tf}
    \end{bmatrix}.
\end{equation}
Variables $l$ and $c_{tf}$ denote the arm length and rotor torque constant, respectively.

\begin{figure}[h!]
    \centering
        \includegraphics[width=0.99\linewidth]{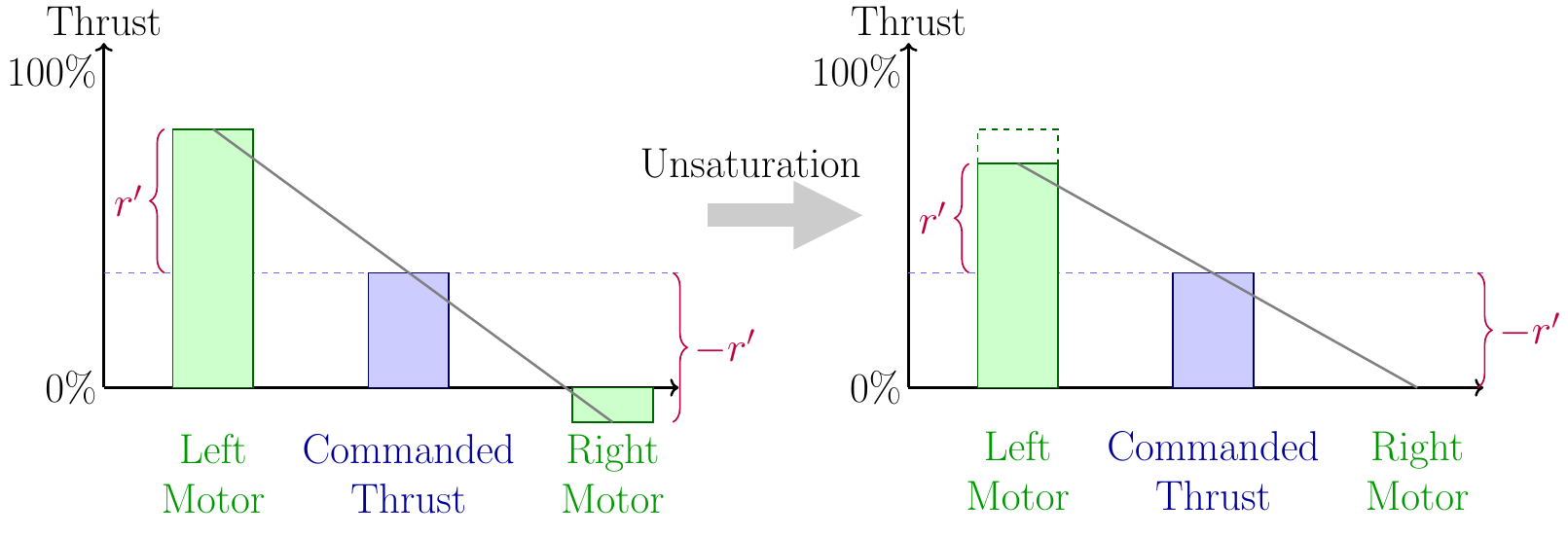}
    \caption{Example of desaturation of two motors by our controller without changing the collective thrust. The image is adapted from~\cite{lorenz_px4_nodate}.}
    \label{fig:px4_unsaturation}
\end{figure}

As mentioned in~\cite{faessler_thrust_2017}, the computed single rotor thrusts must remain within the feasible range $[f_{min}, f_{max}]$. 
This constraint can also be enforced at the throttle level before computing the single rotor thrusts in \eqref{eq:single_rotor_thrusts}.
Our approach uses an iterative motor-mixing strategy at the throttle level.
When actuator saturation occurs, the proposed controller prioritizes preserving collective thrust over torque, giving yaw the least priority.
An example of actuator desaturation is shown in Fig.~\ref{fig:px4_unsaturation}, where single rotor throttles are adjusted while maintaining the collective thrust.

\subsection{PID dynamics for long prediction steps}
\label{sec:flightcontroller}

For longer MPPI prediction timesteps, we adopt the simplified torque computation from~\cite{minarik_model_2024}.
The desired torques are computed as

\begin{equation}
    \bm{\tau}_d =
    \mathbf{J}\frac{\bm{\omega}_c-\bm{\omega}}{\Delta t}
    + \bm{\omega}\times\mathbf{J}\bm{\omega}.
\end{equation}
The corresponding rotor thrusts are obtained using
\begin{equation}
    \bm{f} =
    \bm{\Gamma}^{-1}
    \begin{bmatrix}
        F_t\\
        \bm{\tau}_d
    \end{bmatrix},
\end{equation}
and clipped to the feasible range $[f_{min},f_{max}]$.
The clipped thrusts $\bm{f}_{cl}$ are then converted back to the UAV inputs as
\begin{equation}
    \begin{bmatrix}
        F_t\\
        \bm{\tau}
    \end{bmatrix}
    = \bm{\Gamma}\bm{f}_{cl},
\end{equation}
which is then used to control the UAV model \eqref{eq:uav_eqs}.

\subsection{Model Predictive Path Integral}
\label{sec:mppi}

Trajectory following is a typical task of high-level UAV controllers.
In our architecture, this task is performed by the MPPI.
In the literature, MPPI is typically equipped only with the model of the UAV, i.e., neglecting the effect of the flight controller.
In this paper, we propose to consider both the UAV model and the flight controller in the MPPI.
Therefore, the proposed MPPI control can predict not only the behavior of an ideal UAV, but also take into account the effect of its low-level flight controller.
The remaining model uncertainties and disturbances are compensated by the $\Lone$~adaptive augmentation (Section~\ref{sec:l1_adaptive_controller}).

For this purpose, the state considered in the high-level control is 
$\mathbf{x}= [\mathbf{p},\mathbf{q},\mathbf{v},\bm{\omega},\bm{r},\bm{e}_I]^T$, where
position $\mathbf{p}$, rotation $\mathbf{q}$, velocity $\mathbf{v}$ and body rates $\bm{\omega}$
are described in the UAV model~(\ref{eq:uav_eqs}), 
and normalized motor angular velocities $\bm{r}$ and integral error vector $\bm{e}_I $ of the flight controller are described in (\ref{eq:pid_eqs}).

The task of the MPPI trajectory tracking control is to follow the reference trajectory
represented as a time-parametrized sequence
$\mathcal{T}_{ref}(t) = [\mathbf{p}_r(t), \mathbf{v}_r(t), \mathbf{q}_r(t), \bm{\omega}_r(t)]^T $,
of desired position, velocity, orientation, and body rates, respectively.
MPPI provides control commands for the flight-controller --- 
thrust $F_{MPPI}$ and body-rate $\bm{\omega}_{MPPI}$ (we use subscript $\mppi$ to distinguish them from the $\Lone$ commands).
MPPI assumes that the nominal control sequence
$\bm{u}^{\nom} = (\bm{u}_0^{\nom}, \bm{u}_1^{\nom}, \ldots, \bm{u}_{N-1}^{\nom})$, 
$\bm{u}_i^{\nom} = [F_\mppi, \bm{\omega}_\mppi]^T$
is known from the previous control steps.
The principle of MPPI control is described using the notation and terminology from~\cite{pochobradsky_geometric_2026, minarik_model_2024}.


The current state $\mathbf{x}$ is provided by the UAV state estimator in each iteration.
A set of $K$ disturbance sequences of length $N$, denoted as $\delta\bm{u}^k$, is sampled from a normal distribution with zero mean and covariance matrix $\Sigma$.
Starting from the initial state $\mathbf{x}$, $K$ trajectories (rollouts) are generated by forward simulation of the system dynamics
\begin{equation}
\label{eq:mppi_generation}
\left.
\begin{aligned}
\delta\bm{u}_j^k & \sim \mathcal{N}(0,\Sigma) \\
\bm{u}_j^k &= \bm{u}_j^{\mathrm{nom}} + \delta\bm{u}_j^k \\
\mathbf{x}_{j+1}^k &= \mathbf{x}_j^k + \bm{f}_{\mathrm{RK4}}(\mathbf{x}_j^k, \bm{u}_j^k, \Delta t)
\end{aligned}
\quad
\right\}
\quad
\begin{aligned}
k &= 1,\ldots,K \\
j &= 0,\ldots,N-1 \;.
\end{aligned}
\end{equation}
Function $\bm{f}_{\mathrm{RK4}}$ represents one step of the fourth-order Runge-Kutta method and is used to propagate the UAV dynamics~\eqref{eq:uav_eqs} forward in time, generating the rollouts.
The cost $C^k$ of each rollout is evaluated using a cost function adopted from~\cite{pochobradsky_geometric_2026}.
The cost function reflects how precisely the UAV follows the reference trajectory $\mathcal{T}_{ref}$.
Based on the rollout costs, the weights $w_k$ are given by
\begin{equation}
w_k =
\frac{
\exp\left(-\frac{1}{\lambda}(C^k-\rho)\right)
}{
\sum_{i=1}^{K}\exp\left(-\frac{1}{\lambda}(C^i-\rho)\right)
},
\qquad
\rho=\min_k C^k \;,
\end{equation}
where $\lambda$ determines how strongly the rollout costs affect the computed weights.
Subsequently, the nominal control sequence is updated using a weighted average of the sampled disturbances
\begin{equation}
    \bm{u}_j^{nom} := \sum_{k=1}^K w_k \cdot \bm{u}_j^{k}.
\end{equation}
The first control command from the updated control sequence $\bm{u}_0^\nom = [F_\mppi, \omega_\mppi]^T$ is used by the $\Lone$ controller described in the following subsection.
Its output is added to the MPPI command, and the resulting command is mapped to throttle and sent to the flight controller (see Fig.\ref{fig:control_architecture}).


\subsection{$\Lone$ Adaptive Control}  
\label{sec:l1_adaptive_controller}

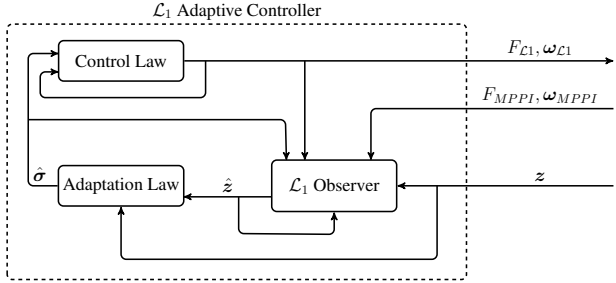
\begin{figure}[ht]
    \centering
    \begin{adjustbox}{width=0.45\textwidth}
        \tikzset{
  >=stealth',
  block/.style={
    rectangle,
    rounded corners,
    draw=black,
    very thick,
    minimum width=3.4cm,
    minimum height=1.1cm,
    align=center,
    font=\Large
  },
  sum/.style={
    circle,
    draw=black,
    very thick,
    minimum size=0.8cm,
    inner sep=1pt,
    font=\Large
  },
  arrow/.style={
    ->,
    very thick,
    rounded corners,
  },
  label/.style={
    font=\Large
  }
}
\begin{tikzpicture}[every node/.append style={font=\Large}]

\node[block] (control) at (2,6) {Control Law};
\node[block, minimum height=1.4cm] (observer) at (7.8,2.6) {$\mathcal{L}_1$ Observer};
\node[block] (adapt) at (2,2.6) {Adaptation Law};

\draw[very thick, dashed, rounded corners] (-1.1,0.05) rectangle (11.4,7.0);
\node[above] at (5.15,7.0) {$\mathcal{L}_1$ Adaptive Controller};

\draw[arrow] (15.4,4.7) -- (8.8,4.7) -- (8.8,3.3);
\node[above] at (13.4,4.7) {$F_{MPPI},\bm{\omega}_{MPPI}$};

\draw[arrow] (4.3,6) -- ++(0,-1.0) -- ++(-4.5,0) -- ++(0,0.7) -- (0.3,5.7);

\draw[arrow] (control.east) -- (15.4,6);
\node[above] at (13.4,6) {${F}_{\mathcal{L}1},\bm{\omega}_{\mathcal{L}1}$};
\draw[arrow] (7,6) -- (7,3.3);

\draw[arrow] (15.4,2.6) -- (observer.east);
\node[above] at (13.4,2.6) {$\bm{z}$};
\draw[arrow] (10.6,2.6) -- (10.6,0.6) -- (2,0.6) -- (adapt.south);

\draw[arrow] (6.1,2.3) -- (3.7,2.3) node[midway, above] {$\hat{\bm{z}}$};
\draw[arrow] (5.2, 2.3) -- ++(0,-1) -- ($(observer.south) + (0, -0.6)$) -- (observer.south);

\draw[arrow] (adapt.west) -- ++(-0.8,0) node[pos=0.6, above] {$\hat{\bm{\sigma}}$} -- ++(0,3.6) -- (0.3,6.2);
\draw[arrow] ($(adapt.west) + (-0.8,1.8)$) -- ++ (7, 0) -- (6.5,3.3);

\end{tikzpicture}
    \end{adjustbox}
    \caption{The architecture of the $\mathcal{L}_1$ adaptive controller is composed of three main blocks: the observer, adaptation law, and control law. Based on the observed state $\hat{\bm{z}}$, uncertainties~$\hat{\bm{\sigma}}=[\hat{\bm{\sigma}}_m, \hat{\bm{\sigma}}_{um}]^T$ are estimated using an adaptation law. The control law then yields the $\mathcal{L}_1$ adaptive controller command  $\bm{u}_{\mathcal{L}1}$  based on the estimated uncertainties.}
    \label{fig:l1_architecture}
\end{figure}

As illustrated in Fig.~\ref{fig:control_architecture}, the $\Lone$ adaptive controller is cascaded with MPPI. 
Whereas MPPI tracks the desired trajectory, the $\Lone$ adaptive controller compensates for model uncertainties and external disturbances.
The inner architecture, consisting of the $\Lone$ observer, adaptation law, and control law, is shown in Fig.~\ref{fig:l1_architecture} and described in this section.

The $\mathcal{L}_{1}$ adaptive controller is implemented according to the nonlinear reference model proposed in~\cite{wang_l1_2012}.
Our approach follows the implementations presented in~\cite{hanover_performance_2022}, but the adaptive command consists of the collective thrust $F_{\mathcal{L}1}$ and desired body rates $\bm{\omega}_{\mathcal{L}1}$.
Both matched and unmatched uncertainties are estimated using a piecewise adaptation law~\cite{li_l1_2012}.
Considering these uncertainties, the UAV dynamics \eqref{eq:uav_eqs} can be rewritten as
\begin{equation}
    \begin{aligned}
        &\dot{\mathbf{v}} = \frac{1}{m} \mathbf{R}(\mathbf{q}) \left(
        \begin{bmatrix}
            0 \\ 0 \\ F_{t}
        \end{bmatrix} + \bm{\varsigma} - \mathbf{D}\mathbf{R}^T(\mathbf{q})\mathbf{v} \right)
        + \mathbf{g} \;,
        \\
        &\dot{\bm{\omega}} =
        \mathbf{J}^{-1}\left(
        \bm{\tau} - \bm{\omega} \times \mathbf{J}\bm{\omega}
        \right) + {\bm{\xi}} \;,
    \end{aligned}
\end{equation}
where $\bm{\varsigma}=[\varsigma_x,\varsigma_y,\varsigma_z]^T$ represents translational force uncertainties and $\bm{\xi}=[\xi_x,\xi_y,\xi_z]^T$ represents body rate uncertainties.
Since a quadrotor can generate linear acceleration only along its body z-axis, the uncertainties $\bm{\sigma}_{um}=[\varsigma_x,\varsigma_y]^T$ are unmatched and cannot be directly compensated through the control input.
The matched uncertainty is defined as $\bm{\sigma}_{m}=[\varsigma_z,\xi_x,\xi_y,\xi_z]^T$, which can be compensated by the $\mathcal{L}_{1}$ adaptive controller.

We define a reduced state $\mathbf{z}=[\mathbf{v},\bm{\omega}]^T$ and express the rotation matrix as
$\mathbf{R}(\mathbf{q})=[\bm{b}_1~\bm{b}_2~\bm{b}_3]$.
The $\mathcal{L}_{1}$ adaptive command is denoted as
$\bm{u}_{\mathcal{L}1}=[F_{\mathcal{L}1},\bm{\omega}_{\mathcal{L}1}]^T$.
The system dynamics can be separated into nominal and uncertain components as
\begin{equation}
    \dot{\bm{z}} =
    \bm{f}(\mathbf{R}(\mathbf{q}))
    + \bm{g}(\mathbf{R}(\mathbf{q}))
    (\bm{u}_{\mathcal{L}1}+\bm{\sigma}_{m})
    + \bm{g}^{\perp}(\mathbf{R}(\mathbf{q}))\bm{\sigma}_{um} \;,
\end{equation}
where $\bm{f}(\mathbf{R}(\mathbf{q}))$ represents the nominal dynamics while $\bm{g}(\mathbf{R}(\mathbf{q}))$ and $\bm{g}^{\perp}(\mathbf{R}(\mathbf{q}))$ are the input distribution matrices.

Since the control input consists of desired body rates rather than direct torque commands, the rotational dynamics are approximated using the first-order relation
$\dot{\bm{\omega}}\approx(\bm{\omega}_{c}-\bm{\omega})/\Delta t$.
This approximation introduces an error that may be interpreted by the $\mathcal{L}_{1}$ controller as an uncertainty term $\bm{\xi}$, potentially causing unnecessary compensation.
However, this effect is considered acceptable if the approximation error remains small compared to the actual system uncertainties.

The nominal dynamics obtained from the MPPI controller are defined as
\begin{equation}
    \bm{f}(\mathbf{R}(\mathbf{q}))=
    \begin{bmatrix}
        \frac{{F}_{MPPI}}{m}\mathbf{b}_3
        -\mathbf{R}(\mathbf{q})\mathbf{D}\mathbf{R}^T(\mathbf{q})\mathbf{v}
        +\mathbf{g}
        \\
        \frac{\bm{\omega}_{MPPI}-\bm{\omega}}{\Delta t}
    \end{bmatrix},
\end{equation}
where ${F}_{MPPI}$ is the collective thrust command and $\bm{\omega}_{MPPI}$ is the commanded body rate, both computed by the MPPI controller at the current iteration.

The matched and unmatched uncertainties enter the system through the aforementioned input distribution matrices $\bm{g}(\mathbf{R}(\mathbf{q}))$ and $\bm{g}^{\perp}(\mathbf{R}(\mathbf{q}))$.
To ensure that the uncertainty $\bm{\xi}$ represents body-rate disturbances, the rotational part of $\bm{g}(\mathbf{R}(\mathbf{q}))$ is defined as $\mathbf{I}/\Delta t$.
This definition ensures that the timestep terms cancel in \eqref{eq:l1_obs_discrete}.
The uncertainty distribution matrices are therefore defined as
\begin{equation}
    \begin{aligned}
        &\bm{g}(\mathbf{R}(\mathbf{q})) =
        \begin{bmatrix}
            \frac{\bm{b}_3}{m} & \mathbf{0}^T
            \\
            \mathbf{0} & \frac{\mathbf{I}}{\Delta t}
        \end{bmatrix},
        &&
        \bm{g}^{\perp}(\mathbf{R}(\mathbf{q})) =
        \begin{bmatrix}
            \frac{\bm{b}_1}{m} & \frac{\bm{b}_2}{m}
            \\
            \mathbf{0} & \mathbf{0}
        \end{bmatrix}.
    \end{aligned}
\end{equation}

The Hurwitz matrix $\mathbf{A}_s$ is defined to determine the observer convergence rate.
The $\mathcal{L}_{1}$ observer (see Fig.~\ref{fig:l1_architecture}) is then formulated as
\[
    \dot{\hat{\bm{z}}} =
    \bm{f}(\mathbf{R}(\mathbf{q}))
    + \bm{g}(\mathbf{R}(\mathbf{q}))
    (\bm{u}_{\mathcal{L}1}+\bm{\sigma}_m)
    + \bm{g}^{\perp}(\mathbf{R}(\mathbf{q}))\bm{\sigma}_{um}
    + \mathbf{A}_s\tilde{\bm{z}} \;,
\]
where $\tilde{\bm{z}}=\hat{\bm{z}}-\bm{z}$ is the state estimation error.
The $\mathcal{L}_{1}$ observer estimates the state $\hat{\bm{z}}$, while the actual state $\bm{z}$ is obtained from the UAV state estimator.
Defining $\bm{\Phi}(iT_s)=\mathbf{A}_s^{-1}(e^{\mathbf{A}_sT_s}-\mathbf{I})$ and 
$\bm{\mu}(iT_s)=e^{\mathbf{A}_sT_s}\tilde{\bm{z}}(iT_s)$, where $T_s$ denotes the adaptation timestep, the piecewise-constant adaptation law (see Fig.~\ref{fig:l1_architecture}) is given by
\begin{equation}
    \begin{bmatrix}
        \hat{\bm{\sigma}}_m(iT_s) \\
        \hat{\bm{\sigma}}_{um}(iT_s)
    \end{bmatrix}
    =
    -\mathbf{I}_{6\times6}
    \mathbf{G}^{-1}(iT_s)
    \bm{\Phi}^{-1}(iT_s)
    \bm{\mu}(iT_s) \;,
\end{equation}
where $\mathbf{G}=[\bm{g}(\mathbf{R}(\mathbf{q})),\bm{g}^{\perp}(\mathbf{R}(\mathbf{q}))]$ and $iT_s$ denotes the $i$-th adaptation step.
The $\mathcal{L}_{1}$ control law (see Fig.~\ref{fig:l1_architecture}) applies a~low-pass filter to the estimated matched uncertainty, resulting in the adaptive command
\begin{equation}
    \label{eq:l1_cmd_filtering}
    \bm{u}_{\mathcal{L}1,k}
    =
    \bm{u}_{\mathcal{L}1,k-1}e^{-\bm{\omega}_{co}T_s}
    -
    \hat{\bm{\sigma}}_{m,k}
    (1-e^{-\bm{\omega}_{co}T_s}) \;,
\end{equation}
where $\bm{\omega}_{co}$ denotes the cutoff frequency of the first-order low-pass filter.
The $\mathcal{L}_{1}$ observer is propagated in discrete time as
\begin{align}
\label{eq:l1_obs_discrete}
\begin{split}
    \hat{\bm{z}}_{k+1}
    =
    \hat{\bm{z}}_k
    &+
    \biggl[
    \bm{f}_k
    +
    \bm{g}_k(\bm{u}_{\mathcal{L}1,k}+\hat{\bm{\sigma}}_{m,k})\\
    &+
    \bm{g}^{\perp}_k\hat{\bm{\sigma}}_{um,k}
    +
    \mathbf{A}_s\tilde{\bm{z}}_k
    \biggr]T_s .
\end{split}
\end{align}

The adaptive command from \eqref{eq:l1_cmd_filtering} is added to the MPPI command as
\begin{equation}
    \bm{u}_d =
    \begin{bmatrix}
        F_t \\
        \bm{\omega}_c
    \end{bmatrix}
    =
    \bm{u}_{MPPI}
    +
    \bm{u}_{\mathcal{L}1,k} \;.
\end{equation}
The desired collective throttle $t_c$ is then obtained from the desired thrust $F_t$.
Together with the desired body rates $\bm{\omega}_c$, it is sent to the low-level flight controller (see Fig.~\ref{fig:control_architecture}).

\subsection{Considering Communication Delay}
\begin{figure}[!h!]
\centering
\includegraphics[width=\linewidth]{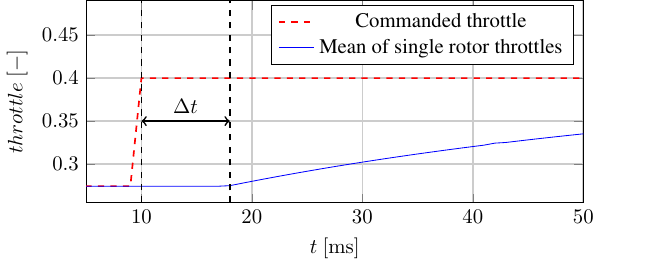}
\vspace{-0.7cm}
\caption{The mean motor throttle response to a~step command, highlighting the delay $\Delta t$ between the applied throttle command and the motor response controlled by the low-level controller.}
\label{fig:cmd_px4_delay}
\end{figure}

Since the MPPI is a high-level controller, the computed command is sent to the low-level flight controller, which controls the rotor actuators. 
This separation introduces a~communication delay between sending the command and the beginning of the rotor throttle response, as illustrated in Fig.~\ref{fig:cmd_px4_delay}.
\vspace{-0.8em}
To compensate for this delay, the initial state of each MPPI iteration is propagated forward using the last applied command $\bm{u}_{last}$
\begin{equation}
    \mathbf{x}_{0} = \mathbf{x} + \bm{f}_{\mathrm{RK4}}(\mathbf{x}, \bm{u}_{last}, \Delta t) \; ,
\end{equation}
where $\Delta t$ is obtained from measurements such as the one shown in Fig.~\ref{fig:cmd_px4_delay}.
The predicted state $\mathbf{x}_{0}$ is then used as the initial state for generating the MPPI rollouts.

\subsection{Separation of $\mathcal{L}_1$ Controller and MPPI} \label{sec:separation}

The MPPI prediction is initialized from the current motor state $\bm{\Omega}$ (or rather its normalized value $\bm{r}$). 
Since the measured motor state already includes the contribution of the $\mathcal{L}_1$ adaptive controller, this contribution must be removed before the MPPI iteration starts.
Otherwise, in the presence of uncertainty, the MPPI prediction assumes that the $\mathcal{L}_1$ command will persist without accounting for the corresponding uncertainty compensation. 
This results in an inconsistent prediction and can cause the MPPI and $\mathcal{L}_1$ controllers to counteract each other.

The single-rotor forces corresponding to the current motor speeds are computed from \eqref{eq:single_rotor_thrusts} as
\begin{equation}
    \bm{F}_\Omega = f_{max}(\bm{r}_\Omega \otimes \bm{r}_\Omega),
\end{equation}
where $\bm{r}_\Omega$ is the vector of normalized motor speeds computed from the measured motor speeds using \eqref{eq:motor_speed_to_throttle}.
The contribution of the MPPI controller is obtained by subtracting the $\mathcal{L}_1$ thrust from the measured forces,
\begin{equation}
    \bm{F}_r = \bm{F}_\Omega - \frac{\bm{F}_{\mathcal{L}1}}{n_{mot}},
\end{equation}
where $n_{mot}$ is the number of motors, $\bm{F}_{\mathcal{L}1} = \mathds{1}F_{\mathcal{L}1}$ and $F_{\mathcal{L}1}$ is the thrust command from the previous iteration.
The corresponding normalized motor speeds are then computed element-wise as
\begin{equation}
    \bm{r} = \sqrt{\frac{\bm{F}_r}{f_{max}}},
\end{equation}
and used to initialize the actuator dynamics described in Sec.~\ref{sec:lol_dynamics}.

\section{Results}
\begin{figure*}[!b]
\centering
\includegraphics[width=0.9\linewidth]{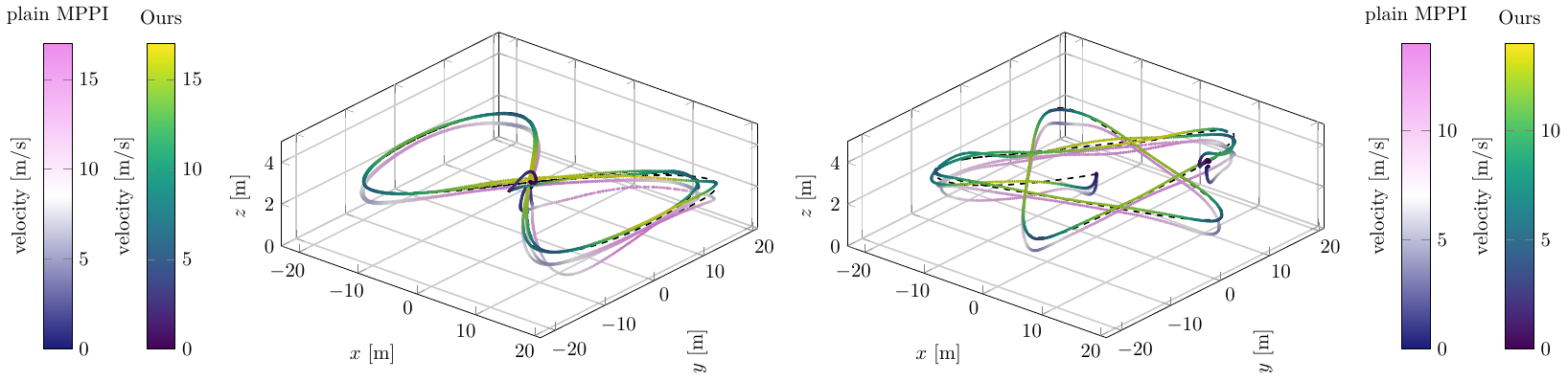}
\caption{Reference and tracked trajectories under an additional payload using the proposed $\mathcal{L}_1$-MPPI controller and plain MPPI.}
\label{fig:heatmap_payload_trl1}
\end{figure*}
This section presents the experimental design and results demonstrating the trajectory tracking and adaptive capabilities of the proposed $\mathcal{L}_1$-MPPI controller. 
The proposed approach is compared with existing methods to assess its performance.
Additionally, an ablation study evaluates the contribution of each feature introduced in Sec.~\ref{sec:methodology}.

Simulation experiments were performed on a~laptop equipped with a~12-core Intel Core i5-12450H CPU and an NVIDIA GeForce RTX 3050 GPU. 
Real-world experiments used a~Jetson Orin NX onboard computer equipped with an 8-core Arm Cortex-A78AE CPU and an Ampere GPU. 
The controller outputs desired collective throttle and body-rate commands, which are tracked by a PX4 low-level flight controller. 
Reference trajectories were generated offline using a~polynomial trajectory planner~\cite{wang_generating_2022} and provided to the \mbox{$\mathcal{L}_1$-MPPI} controller during flight.


\subsection{Designed Experiments} \label{sec:designed_exp}
The proposed \mbox{$\mathcal{L}_1$-MPPI} controller is compared with the plain MPPI implementation from~\cite{pochobradsky_geometric_2026}, denoted as \textit{MPPI (plain)}, and with our MPPI implementation augmented by a mass estimator, denoted as \textit{MPPI (mass est.)}.
An ablation study further evaluates the contribution of the proposed components.

The evaluated \mbox{$\mathcal{L}_1$-MPPI} configurations are:
\begin{itemize}
    \item \textit{Ours} -- Our proposed $\mathcal{L}_1$-MPPI.
    \item \textit{Ours (no delay)} -- Our method without communication delay compensation.
    \item \textit{Ours ($\mathcal{L}_1$ tr. only)} -- $\mathcal{L}_1$ Our method with augmentation applied only to the translational dynamics.
    \item \textit{Ours (no $\mathcal{L}_1$)} -- Our method without $\mathcal{L}_1$ augmentation.
    \item \textit{Ours (no PID)} -- Our method without integration of low-level controller and motor dynamics models.
\end{itemize}

Two trajectories shown in Fig.~\ref{fig:heatmap_payload_trl1} are selected to evaluate agile UAV flight. 
They are generated using a polynomial trajectory planner~\cite{wang_generating_2022}.
Both trajectories reach accelerations of up to \SI{3.5}{\g}, with maximum velocities of \SI{17.84}{\meter\per\second} and \SI{15.21}{\meter\per\second} for the \textit{figure 8} and \textit{hypotrochoid} trajectories, respectively.
The \textit{figure 8} trajectory evaluates high-speed flight with curved sections, while the \textit{hypotrochoid} trajectory tests tracking performance during sharp turns and rapidly varying accelerations.

To evaluate the adaptive capabilities of the proposed approach, three configurations are considered for each trajectory: the nominal model (\textit{No change}), an unknown additional payload that increases the UAV's mass by \SI{35}{\percent} (\textit{Payload}), and doubled drag coefficients $c_x$, $c_y$, and $c_z$ (\textit{Drag}). 
These scenarios evaluate the effects of low-level dynamic modeling, $\mathcal{L}_1$ augmentation, and adaptation to mass and aerodynamic uncertainties.

\subsection{Simulated Results}

The experiments described in Sec.~\ref{sec:designed_exp} were conducted in simulation. 
Each scenario was repeated at least 10 times, resulting in over 400 simulated flights.
Tracking performance was evaluated using the positional Root Mean Square Error (RMSE).
The results are summarized in Tab.~\ref{tab:tracking_results}.

\begin{table}[t]
\centering
\caption{Results of the tracking experiments.}
\label{tab:tracking_results}
\resizebox{\columnwidth}{!}{%
\begin{tabular}{c | l |
                c c |
                c c}
\toprule
\multirow{2}{*}{\textbf{Traj.}} &
\multirow{2}{*}{\textbf{Controller}} &
\multicolumn{2}{c|}{\textbf{Figure 8}} &
\multicolumn{2}{c}{\textbf{Hypotrochoid}} \\
& & \textbf{pos. RMSE [m]} & \textbf{Gain [\%]} & \textbf{pos. RMSE [m]} & \textbf{Gain [\%]} \\
\midrule
\multirow{7}{*}{\rotatebox{90}{\textit{No change}}} 
& MPPI (plain)                    & $0.673 \pm 0.025$ & --      & $0.609 \pm 0.021$ & --      \\
& Ours                            & $0.529 \pm 0.026$ & $21.35$ & $0.437 \pm 0.024$ & $28.20$ \\
& Ours (no delay)                 & $0.529 \pm 0.023$ & $21.36$ & $0.439 \pm 0.010$ & $27.90$ \\
& Ours ($\mathcal{L}_1$ tr. only) & $\bm{0.509 \pm 0.019}$ & $\bm{24.35}$ & $0.451 \pm 0.011$ & $25.85$ \\
& Ours (no $\mathcal{L}_1$)       & $0.524 \pm 0.015$ & $22.16$ & $0.439 \pm 0.014$ & $27.86$ \\
& Ours (no PID)                   & $0.684 \pm 0.023$ & $-1.60$ & $0.644 \pm 0.031$ & $-5.76$ \\
& MPPI (mass est.)                & $0.523 \pm 0.022$ & $22.26$ & $\bm{0.437 \pm 0.018}$ & $\bm{28.23}$ \\
\midrule
\multirow{7}{*}{\rotatebox{90}{\textit{Payload}}}
& MPPI (plain)                    & $0.750 \pm 0.060$ & --      & $0.657 \pm 0.028$ & --      \\
& Ours                            & $\bm{0.330 \pm 0.034}$ & $\bm{56.07}$ & $\bm{0.256 \pm 0.032}$ & $\bm{61.14}$ \\
& Ours (no delay)                 & $0.413 \pm 0.063$ & $44.95$ & $0.281 \pm 0.035$ & $57.22$ \\
& Ours ($\mathcal{L}_1$ tr. only) & $0.419 \pm 0.082$ & $44.18$ & $0.349 \pm 0.039$ & $46.99$ \\
& Ours (no $\mathcal{L}_1$)       & $0.661 \pm 0.047$ & $11.93$ & $0.535 \pm 0.022$ & $18.57$ \\
& Ours (no PID)                   & $0.450 \pm 0.167$ & $39.95$ & $0.376 \pm 0.007$ & $42.85$ \\
& MPPI (mass est.)                & $0.517 \pm 0.020$ & $31.06$ & $0.434 \pm 0.014$ & $33.95$ \\
\midrule
\multirow{7}{*}{\rotatebox{90}{\textit{Drag}}}
& MPPI (plain)                    & $1.038 \pm 0.037$ & --      & $0.776 \pm 0.041$ & --      \\
& Ours                            & $0.715 \pm 0.096$ & $31.12$ & $0.550 \pm 0.066$ & $29.12$ \\
& Ours (no delay)                 & $0.753 \pm 0.078$ & $27.46$ & $0.594 \pm 0.060$ & $23.39$ \\
& Ours ($\mathcal{L}_1$ tr. only) & $0.832 \pm 0.100$ & $19.87$ & $0.621 \pm 0.043$ & $19.94$ \\
& Ours (no $\mathcal{L}_1$)       & $1.172 \pm 0.084$ & $-12.89$ & $0.814 \pm 0.007$ & $-4.98$ \\
& Ours (no PID)                   & $\bm{0.525 \pm 0.075}$ & $\bm{49.40}$ & $\bm{0.460 \pm 0.058}$ & $\bm{40.67}$ \\
& MPPI (mass est.)                & $1.622 \pm 0.105$ & $-56.20$ & $1.047 \pm 0.063$ & $-34.94$ \\
\bottomrule
\end{tabular}
}
\end{table}
The proposed $\mathcal{L}_1$-MPPI controller improves tracking performance over plain MPPI in all tested conditions, with an average improvement of approximately \SI{24.78}{\percent} under nominal conditions, increasing to \SI{58.61}{\percent} with an additional payload and \SI{30.12}{\percent} under increased drag.
Compared with \textit{Ours (no $\mathcal{L}_1$)}, the $\mathcal{L}_1$ augmentation has an insignificant effect on nominal tracking, while it substantially improves performance under parameter mismatches.

Accounting for the low-level PID dynamics has a clear impact on tracking performance.
Removing the PID dynamics (\textit{Ours (no PID)}) degrades performance on average by approximately \SI{25}{\percent} under nominal conditions and \SI{16}{\percent} with an additional payload.
Interestingly, the opposite trend is observed under drag mismatch, where neglecting the PID dynamics improves tracking by approximately \SI{17}{\percent}.
A possible explanation is that the $\mathcal{L}_1$ compensation of the drag disturbance interacts less favorably with the additional low-level controller states in the prediction model, as the drag force depends on the attitude, which is highly influenced by the low-level controller.

Finally, the mass estimator performs comparably to the proposed approach under nominal conditions and with an additional payload, but its performance degrades substantially under increased drag, resulting in approximately \SI{67}{\percent} higher RMSE than the proposed approach.
This suggests that $\mathcal{L}_1$ adaptation provides a more general mechanism for compensating for uncertainties that cannot be represented by a mass change alone.

\subsection{Real-world Experiments}
We chose three scenarios to track a figure 8 trajectory in the real-world experiment, achieving velocities up to \SI{13.51}{\meter\per\second} and accelerations reaching \SI{2.5}{\g}.
First, all model parameters are unchanged (\textit{No change}).
Second, the \textit{Mass mismatch} condition was created by internally reducing the modeled UAV mass from \SI{1.43}{\kilogram} to \SI{0.8}{\kilogram}, corresponding to a \SI{79}{\percent} apparent mass increase for the controller, without adding a~physical payload.
Finally, a~physical payload is attached, which increases the UAV mass by \SI{35}{\percent} (\textit{Real payload}) and also influences the inertia.
The real-world experiments are summarized in Tab.~\ref{tab:real_tracking_results_fig8}.

\begin{table}[!t]
\centering
\caption{RMSE results of tracking figure 8 in the real world.}
\label{tab:real_tracking_results_fig8}
\resizebox{\columnwidth}{!}{%
\begin{tabular}{l|ccc}
\toprule
\textbf{Condition} &
\textbf{Ours [m]} &
\textbf{Ours ($\mathcal{L}_1$ tr. only) [m]} &
\textbf{Ours (no $\mathcal{L}_1$) [m]} \\
\midrule
No change
    & $0.703 $
    & $0.707 $
    & $\bm{0.654}$ \\
Mass mismatch
    & $\bm{0.727}$
    & $0.981 $
    & $1.266 $ \\
Real payload
    & $\bm{0.621 }$
    & $0.699 $
    & -- \\
\bottomrule
\end{tabular}
}
\end{table}

\begin{figure}[!h]
    \vspace{-0.5em}
    \centering
    \includegraphics[width=\columnwidth]{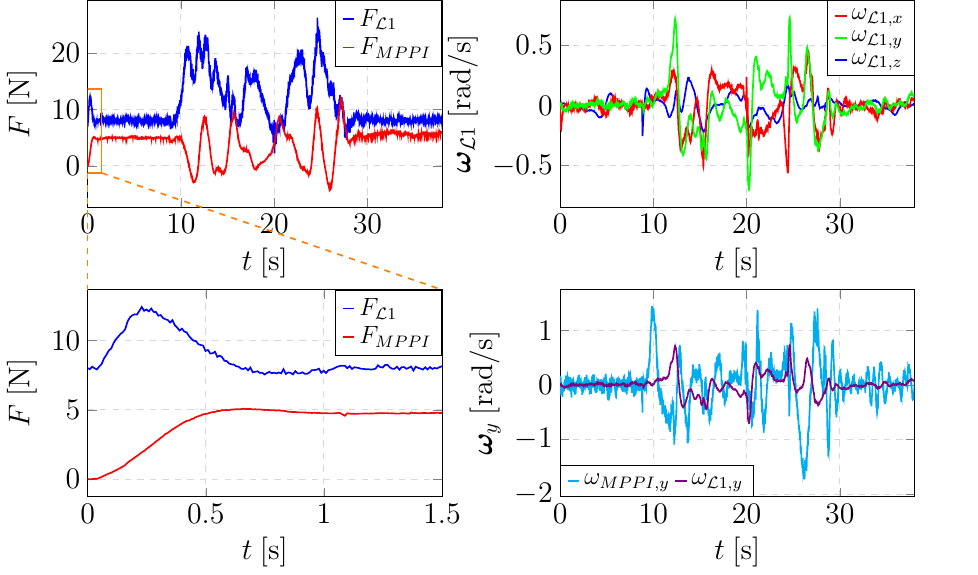}
    \caption{Commands from the MPPI and $\mathcal{L}_1$ adaptive controller during the real-world figure 8 tracking experiment. The force plots show the collective thrust of each controller over the entire trajectory and its beginning. The top-right plot illustrates the rotational dynamics compensation by the $\mathcal{L}_1$ adaptive controller. Bottom-right plot compares the roll rate command of MPPI and $\mathcal{L}_1$ adaptive controller.}
    \label{fig:l1_cmd_fig8}
\end{figure}

The $\mathcal{L}_1$ augmentation of the proposed controller shows a~similar impact in real-world experiments to the simulated ones. 
$\mathcal{L}_1$-MPPI increases the position RMSE by \SI{7.49}{\percent} in \textit{No change} scenario comapred to \textit{Ours (No $\mathcal{L}_1$)}. 
However, in the presence of additional payload, full $\mathcal{L}_1$ augmentation demonstrates its capability of compensating for the unknown payload, even outperforming \textit{Ours (No $\mathcal{L}_1$)} in nominal settings, similarly to simulations.
This may be due to the MPPI controller's cost function, which was not tuned to the new dynamics.
Fig.~\ref{fig:l1_cmd_fig8} illustrates the commands computed by the $\mathcal{L}_1$-MPPI in the real-world \textit{Mass mismatch} scenario.

\section{Conclusion}
This work proposed an $\mathcal{L}_1$-MPPI controller for agile flight in the presence of model uncertainties and external disturbances. 
To the best of our knowledge, it is the first MPPI-based controller that incorporates a low-level controller and motor dynamics.
Additionally, $\mathcal{L}_1$-MPPI is the first adaptive MPPI suitable for onboard control.
We evaluated the trajectory tracking performance in simulation at speeds reaching \SI{17.84}{\meter\per\s} and accelerations up to \SI{3.5}{\g}.
The proposed method demonstrates its tracking performance across various scenarios, outperforming both the existing MPPI approach on agile trajectories and the mass estimation approach in flight under uncertainties.
Real-world experiments demonstrate the adaptability of the $\mathcal{L}_1$-MPPI without performance degradation while achieving speeds up to \SI{13.5}{\meter\per\s} and accelerations up to \SI{2.5}{\g}. 


\bibliographystyle{IEEEtran}
\bibliography{references}

\end{document}